# Identification of Pediatric Sepsis Subphenotypes for Enhanced Machine Learning Predictive Performance: A Latent Profile Analysis

Tom Velez PhD[1*], Tony Wang PhD[2], Ioannis Koutroulis MD[3], James Chamberlain MD[3], Amit Uppal MD[4], Seife Yohannes MD[5], Tim Tschampel[2]; Emilia Apostolova PhD[6];

[1]Computer Technology Associates, Cardiff, CA; [2]Imedacs, Ann Arbor, MI; [3] Emergency Medicine, Children's National Health System, Washington, DC, [4]Critical Care Medicine, NYU Langone Health, New York, [5]MedStar Health System, Washington, DC, [6]Language.ai, Chicago, IL

*Corresponding author: tom.velez@cta.com

## Abstract

**Background:** While machine learning-based models are rapidly emerging as promising screening tools in critical care medicine, the identification of homogeneous subphenotypes within populations with heterogeneous conditions such as pediatric sepsis may facilitate attainment of high-predictive performance of these prognostic algorithms. This study is aimed to identify subphenotypes of pediatric sepsis and demonstrate the potential value of partitioned data/subtyping in predictive analytics.

**Methods:** This was a retrospective study of clinical data extracted from medical records of 6,446 pediatric patients that presented to emergency departments and were admitted at a major hospital system in the Washington DC area. Vitals and labs associated with patients meeting the diagnostic criteria for sepsis in this cohort were used to perform latent profile analysis. Gradient boosted machine and random forest algorithms were used to explore the predictive performance benefits of reduced training data heterogeneity via label profiling.

**Results:** In total 134 (2.1%) patients met the diagnostic criteria for sepsis in this cohort and latent profile analysis identified four profiles/subphenotypes of pediatric sepsis. Profiles 1 and 3 had the lowest mortality and included pediatric patients from different age groups. Profile 2 patients were characterized by respiratory dysfunction while profile 4 patients characterized by neurological dysfunction (lowest total Glasgow Coma Score) and highest mortality rate (22.2%). Multiple machine learning experiments comparing the predictive performance of models derived without training data profiling against profile targeted models suggest statistically significant improved performance of prediction can be obtained. For example, area under ROC curve (AUC) obtained to predict profile 4 with 24-hour data (AUC = .998, $p < .0001$) compared favorably with the AUC obtained from the model considering all profiles as a single homogeneous group (AUC = .918) with 24-hour data.

**Conclusion:** This study utilized LPA to identify four clinically meaningful pediatric sepsis subphenotypes in training data and, for two of these subclasses with the highest mortality, derived statistically significant enhanced predictive models on the partitioned data. These experiments suggest that LPA is an applicable tool useful in analyzing heterogeneous pediatric sepsis to identify subphenotypes beneficial for building enhanced predictive pediatric sepsis models. Additional studies with larger data samples are needed to validate our findings.

## 1 Introduction

The early recognition and management of sepsis remain among the greatest challenges in pediatric emergency medicine [1]–[3]. Sepsis is directly responsible for more than 4,000 childhood deaths per year in the U.S. and globally is associated with more than 6 million neonatal and early childhood deaths [4]–[6]. One-third of children who die in pediatric intensive care units (PICUs) within the U.S. have severe sepsis [7]. Excluding newborn and maternal stays, in 2012 there were approximately 1.8 million pediatric hospital admissions in the U.S., with an aggregated hospital annual cost of more than $20B [8]. Over 70% of these hospitalizations occurred in general hospitals as opposed to freestanding children's hospitals. [9]

Early recognition and timely, aggressive therapy are of pivotal importance to improve the outcomes of pediatric sepsis patients [1]–[5]. However the automated discrimination of the critically ill from normal pediatric patients presenting with abnormal temperature in the emergency department (ED) is challenging for several reasons [14][15]. Current pediatric sepsis screening tools are largely based on systemic inflammatory response syndrome (SIRS) criteria [16], designed to maximize sensitivity, while exhibiting poor specificity [17] and positive predictive value (PPV) [18], prompting modifications to improve specificity and PPV [19]. Although the recent consensus-based redefinition of adult sepsis (Sepsis-3) [20] selected the sequential organ failure assessment (SOFA) scoring system to address poor SIRS specificity, the derivation and validation of age-adjusted SOFA based screening tools for use in pediatrics [21] remain incomplete. Supervised machine learning (ML) sepsis predictive algorithms with the potential to improve the screening performance of rule-based systems have been proposed [22] [23], while generally limited by lack of large samples of high quality training data [23], i.e. accurately identified sepsis. Moreover, sepsis is highly heterogeneous, with variable clinical presentations depending on the initial site of infection, the stage of sepsis at presentation, causative organism, pattern of acute organ dysfunction (OD), and underlying health status of the patient [24]. Such heterogeneity in population implies heterogeneity in relationships between explanatory and response variables within partitions, potentially posing serious challenges in predictive model building seeking to identify common explanatory data patterns in observed data associated with an outcome [25].

The basic concept of precision medicine is the identification of subphenotypes of patients based on characteristics such as medical history, genetic makeup or electronic health record data (EHR), that will respond to subgroup-targeted personalized drugs or treatments [26]. Many efforts have been made to identify sepsis subphenotypes by using genomics and transcriptomics [27]. While genotyping is not currently routinely performed in daily clinical practice, recent large cohort studies indicate that unsupervised clustering analytics using routinely available clinical data extracted from EHRs may be useful to identify clinically useful partitions towards reducing heterogeneity in syndromic diseases. Specifically, well-validated mixture modeling statistical techniques such as Latent Class Analysis (LCA) [28] and Latent Profile Analysis (LPA) [29]-[30] that aim to recover hidden groups within observed data have been used with readily available clinical data to detect potentially useful subphenotypes associated with different mortality outcomes and response to treatment in adult sepsis [31] [32] and acute respiratory distress syndrome (ARDS) [33].

To our knowledge, current ML algorithms used to predict sepsis do not yet target homogeneous subphenotypes in sepsis populations, and in general ignore most of existing sepsis domain knowledge [34]. The predictive performance benefits of incorporating prior knowledge such as subphenotype diversity into modern ML algorithms are not well understood [35].

In this study, we integrate prior knowledge of the heterogeneity in a pediatric sepsis population into predictive model building by identifying subphenotypes that share common underlying pathophysiology as statistically expressed in observed clinical data (e.g. vitals signs, laboratory measurements). Specifically, we hypothesize that using LPA-identified subphenotypes of sepsis subjects [32] in training data to build predictive models on more homogeneous partitioned data [25] will result in improved predictive validity when compared to models derived from treating all patients as a single homogeneous group. Prior work has demonstrated the value of problem domain classification semantics in the enhancement of mortality risk predictive models [36]. For pediatric sepsis LPA and predictive modeling, following IRB approval, we retrospectively analyzed data abstracted from medical records of 6,446 pediatric patients presenting to one of six MedStar Health System admitting facilities in the DC-Baltimore metropolitan area that occurred from 2013-2018. The median age of this cohort was 9 years with an average length of stay (LOS) of 85.49 hrs.

## 2 Methods and Results: LPA

### 2.1 Data for Latent Profile Analysis
For this study, sepsis was defined as a life-threatening organ dysfunction caused by a dysregulated host response to infection per Sepsis-3 definition [20]. To identify sepsis cases we adopted the method used by Zhang [32], and screened for patients with documented or suspected infection, plus the presence of organ dysfunction. ICD diagnosis codes for a bacterial or fungal infection were used to define infection (see Table

A3 in Appendix for SQL used to detect ICD-9 infection codes Zhang [32]; ICD-10 equivalents were used for post-2015 encounters). Similarly ICD diagnosis and/or procedure codes (identifying mechanically ventilated patients) were used to identify organ dysfunction [32]. A patient was defined to have organ dysfunction if he or she had ICD-9 code (or ICD-10 equivalents) as follows: unspecified thrombocytopenia (287.5), hypotension (458.9), acute and subacute necrosis of liver (570), acute kidney failure (584.9), anoxic brain damage (348.1), shock without mention of trauma (785.59), encephalopathy (348.30), transient mental disorders due to conditions classified elsewhere (293.9), secondary thrombocytopenia (287.49), other and unspecified coagulation defects (286.9), defibrination syndrome (286.6), and hepatic infarction (573.4). If mechanical ventilation (procedures ICD code: 96.70, 96.71, 96.72) was required, it was also defined as organ dysfunction.

Using the definition, 134 (2.1%) of the 6446 pediatric patients meet the diagnostic criteria for sepsis. In seeking pediatric sepsis subphenotypes, we sought to use clinical features following onset in our LPA analysis. The determination of time points of sepsis onset for sepsis patients was based on age-adjusted SOFA criteria [21], shown as a concept map in Fig. 1. Onset time was defined as the time when any component in age-adjusted SOFA was found abnormal.

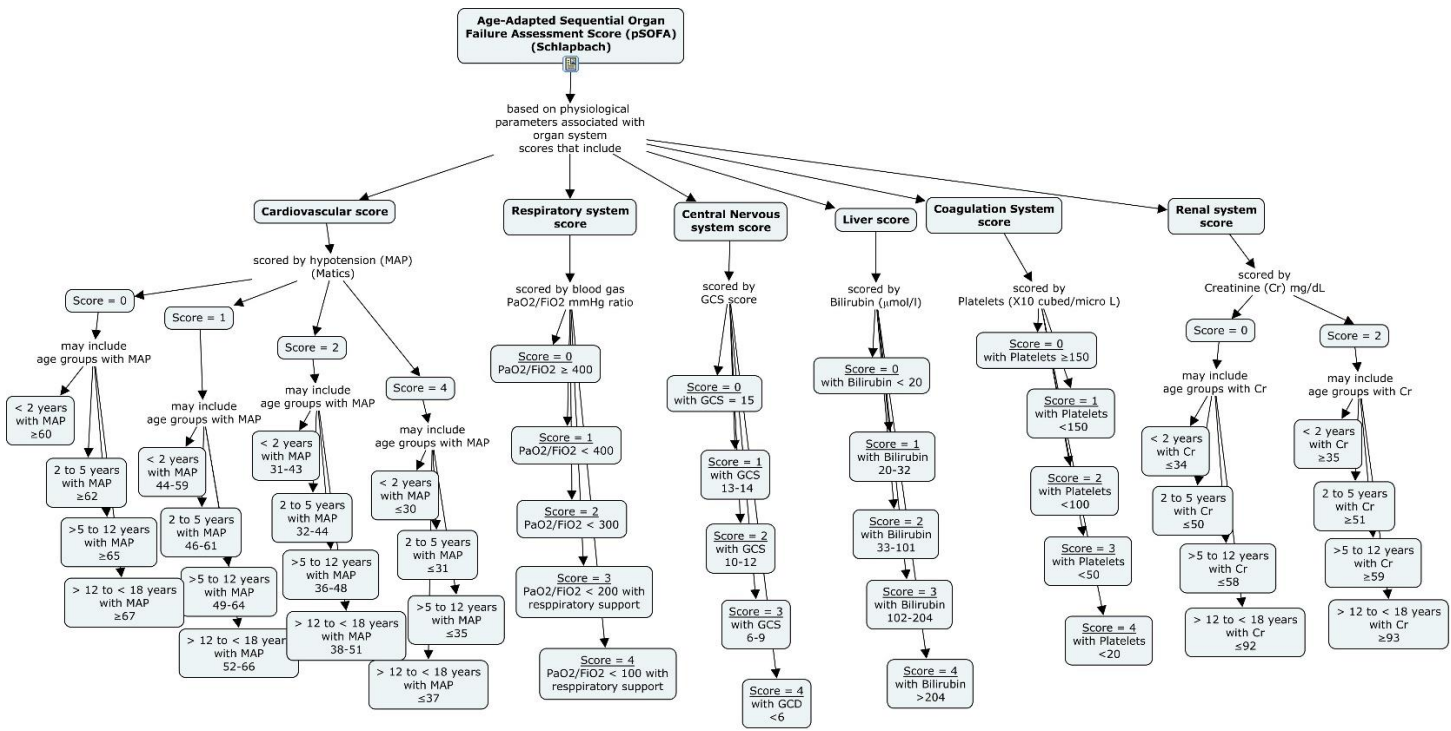

Figure 1 Sepsis-3 criteria concept map: Age-adjusted SOFA score

## 2.2 Latent Profile Analysis

All available post-onset clinical data and lab measurements of 134 sepsis patients were considered as profile-defining variables in the LPA modelling; profiling was conducted without consideration of clinical outcomes. Details on clinical variable selection, data cleaning and a complete list of the clinical variables included in the LPA models are listed in Table 1. Unless specified, medians of measurements were extracted. Median imputation was applied to each variable with missing values. Given the small sample size we were unable to exclude features with large amounts of missing data. To select a model fitting the data best, a series of Latent profile models with different number of components are fitted, and Bayesian Information Criteria (BIC) is used for model selection [37].

Latent profile model estimation is based on Gaussian finite mixture modelling methods [38]. It assumes that the population is composed of a finite number of components. Mixture model parameters, i.e. components' means, covariance structure, and mixing weights, are obtained via the expectation maximization (EM) algorithm. Before LPA modeling, Yeo-Johnson power transformation [39] is applied to ensure approximate normality of continuous variables.

| Variable | %missing | Mean | SD | Min | Median | Max |
| --- | --- | --- | --- | --- | --- | --- |
| Age, yrs. | 0.0 | 10.3 | 7.2 | 0.0 | 11.5 | 18.0 |
| Bicarbonate, mmol/L | 3.7 | 21.1 | 5.5 | 5.0 | 22.0 | 44.0 |
| Bilirubin, mg/dL | 17.2 | 3.0 | 6.9 | 0.1 | 0.6 | 39.3 |
| Chloride, mmol/L | 3.7 | 106.7 | 6.6 | 88.0 | 106.0 | 128.0 |
| Creatinine, mg/dL | 3.7 | 0.8 | 0.5 | 0.2 | 0.7 | 3.7 |
| GCS | 46.3 | 11.2 | 4.5 | 3.0 | 14.0 | 15.0 |
| Eye Opening Response, GCS | 46.3 | 2.8 | 1.3 | 1.0 | 3.0 | 4.0 |
| Best Motor Response, GCS | 46.3 | 4.8 | 1.8 | 1.0 | 6.0 | 6.0 |
| Best Verbal Response, GCS | 46.3 | 3.5 | 1.9 | 1.0 | 5.0 | 5.0 |
| Glucose mg/dL | 79.1 | 164.8 | 107.1 | 70.0 | 125.0 | 600.0 |
| Hematocrit, % | 3.0 | 34.4 | 6.7 | 15.2 | 34.5 | 51.0 |
| Heart rate, /min | 9.0 | 119.5 | 29.9 | 50.0 | 119.5 | 195.0 |
| INR | 35.1 | 1.6 | 0.9 | 0.9 | 1.3 | 5.8 |
| MAP, mmHg | 2.2 | 77.6 | 17.2 | 43.7 | 78.3 | 131.0 |
| PaCO2, mmHg | 64.9 | 39.4 | 11.2 | 10.0 | 39.0 | 71.0 |
| PaO2, mmHg | 65.7 | 142.8 | 107.8 | 24.0 | 111.0 | 443.0 |
| pH | 57.5 | 7.3 | 0.1 | 6.9 | 7.4 | 7.7 |
| Platelet, x10$^9$/L | 3.0 | 224.9 | 150.5 | 8.0 | 210.5 | 821.0 |
| Potassium, mmol/L | 3.7 | 4.1 | 0.8 | 2.6 | 4.0 | 6.9 |
| PTT, s | 50.0 | 38.5 | 14.2 | 23.5 | 34.3 | 89.8 |
| Respiratory rate, /min | 3.0 | 26.4 | 11.2 | 14.0 | 24.0 | 72.0 |
| Sodium, mmol/L | 3.7 | 139.1 | 5.7 | 122.0 | 139.0 | 154.0 |
| Temperature, °C | 23.9 | 37.4 | 1.1 | 35.7 | 37.0 | 40.0 |

**Table 1. Basic statistics of variable used in LPA**

## 2.3 LPA Results

BIC criterion suggests a 4-component VEI model (i.e. diagonal, varying volume, and equal shape) fits the data best (Figure 2). Numbers of patients assigned into subphenotypes are 33 in subphenotype 1, 35 in subphenotype 2, 39 in subphenotype 3, and 27 in subphenotype 4.

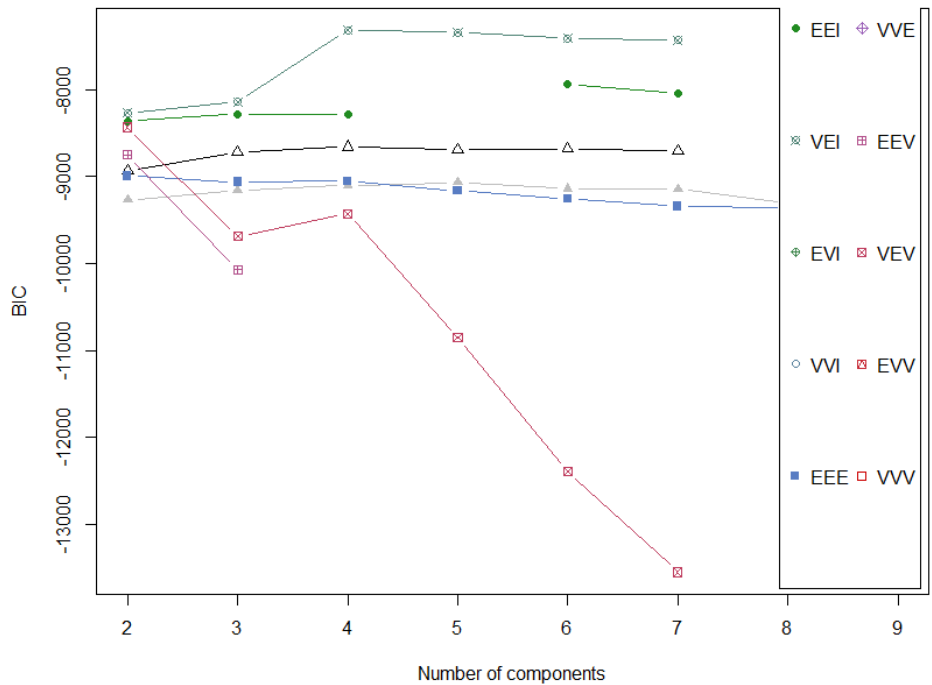

**Figure 2** Models, number of profiles, and corresponding BIC. (See Appendix Table A1 for more information about model indices in legend).

Table 2 and Fig. 3 (box plots of distribution of clinical features) shows differences in clinical features between profiles. Profiles 1 and 3 had the lowest mortality and included pediatric patients from different age groups. Profile 4 patients characterized by neurological dysfunction (lowest GCS total score) and highest mortality rate (22.2%); profile 2 characterized by respiratory dysfunction (low PaO2). Both profiles 2 and 4 patients had significant more vasopressor use, while profile 2 patients had lowest mean blood pressure.

| Variable, mean(SD)/median(IQR) | Subphenotype 1 | subphenotype 2 | Subphenotype 3 | Subphenotype 4 | p |
|---|---|---|---|---|---|
| n | 33 | 35 | 39 | 27 | |
| Age yrs. | 15.48 (3.03) | 8.91 (7.76) | 3.44 (3.12) | 15.67 (4.65) | <0.001 |
| Platelet, x10$^9$/L | 199.50 [144.75, 274.25] | 177.00 [73.50, 329.00] | 211.50 [116.75, 318.50] | 216.00 [100.00, 288.00] | 0.859 |
| PTT, s | 35.40 [32.70, 37.00] | 36.40 [33.50, 47.20] | 33.00 [29.62, 36.75] | 30.30 [26.85, 37.95] | 0.242 |
| INR | 1.20 [1.10, 1.30] | 1.40 [1.30, 2.00] | 1.20 [1.10, 1.30] | 1.30 [1.20, 1.70] | 0.003 |
| Creatinine* | -0.10 (0.13) | -0.24 (0.35) | -0.40 (0.24) | -0.06 (0.21) | <0.001 |
| PaCO2, mmHg | 37.50 (2.12) | 36.80 (12.98) | 59.00 (NA) | 40.87 (9.25) | 0.470 |
| PaO2* | 1.96 (0.31) | 1.93 (0.33) | 2.25 (NA) | 2.15 (0.27) | 0.060 |
| Mean BP, mmHg | 78.19 (12.83) | 74.76 (20.98) | 75.65 (14.53) | 83.85 (19.44) | 0.189 |
| Chloride, mmol/L | 107.26 (4.53) | 105.43 (8.78) | 106.19 (5.37) | 108.44 (6.82) | 0.311 |
| pH | 7.35 (0.05) | 7.30 (0.18) | 7.35 (0.10) | 7.34 (0.16) | 0.746 |
| Bicarbonate, mmol/L | 22.58 (2.88) | 19.63 (7.98) | 21.39 (4.74) | 20.81 (4.23) | 0.172 |
| Hematocrit, % | 34.06 (5.59) | 33.83 (8.22) | 33.03 (5.44) | 37.36 (6.65) | 0.065 |
| Temperature, ºC | 37.58 (1.11) | 37.33 (1.06) | 37.30 (1.07) | 37.20 (1.04) | 0.596 |
| Glucose, mg/dL | 102.50 (4.95) | 193.53 (133.64) | 136.25 (41.96) | 137.43 (65.11) | 0.498 |
| Sodium, mmol/L | 139.16 (3.80) | 137.91 (7.58) | 137.81 (4.65) | 142.41 (5.03) | 0.005 |
| Potassium, mmol/L | 3.86 (0.50) | 4.19 (0.89) | 4.39 (0.96) | 3.98 (0.73) | 0.045 |
| HR, /min | 98.73 (23.70) | 133.38 (28.01) | 128.36 (25.97) | 113.46 (30.18) | <0.001 |
| RR, /min | 18.00 [18.00, 20.00] | 30.00 [20.00, 39.75] | 28.00 [24.00, 34.00] | 17.50 [16.00, 28.25] | <0.001 |
| Bilirubin, mg/dL | 0.60 [0.40, 0.80] | 0.70 [0.40, 3.65] | 0.50 [0.30, 1.20] | 0.50 [0.40, 0.80] | 0.416 |
| Minimum GCS score | 15.00 [15.00, 15.00] | 14.00 [14.00, 15.00] | 15.00 [14.00, 15.00] | 6.00 [3.00, 9.00] | <0.001 |
| GCS, min motor score | 6.00 [6.00, 6.00] | 6.00 [6.00, 6.00] | 6.00 [6.00, 6.00] | 4.00 [1.00, 5.00] | <0.001 |
| GCS, min verbal score | 5.00 [5.00, 5.00] | 5.00 [5.00, 5.00] | 5.00 [5.00, 5.00] | 1.00 [1.00, 1.00] | <0.001 |
| GCS, min eye score | 4.00 [4.00, 4.00] | 3.00 [3.00, 4.00] | 4.00 [4.00, 4.00] | 1.00 [1.00, 2.50] | <0.001 |
| Dopamine, n (%) | 0 (0.0) | 8 (22.9) | 0 ( 0.0) | 4 (14.8) | <0.001 |
| Epinephrine, n (%) | 2 (6.1) | 13 (37.1) | 9 (23.1) | 9 (33.3) | 0.01 |
| Phenylephrine, n (%) | 0 (0.0) | 3 ( 8.6) | 0 ( 0.0) | 9 (33.3) | <0.001 |
| Vasopressin, n (%) | 0 (0.0) | 2 ( 5.7) | 0 ( 0.0) | 4 (14.8) | 0.009 |
| Death, n (%) | 0 (0.0) | 2 ( 5.7) | 0 ( 0.0) | 6 (22.2) | 0.001 |

**Table 2 Simple Statistics between 4 subphenotypes** Abbreviations: PTT partial thrombin time, BP blood pressure, GCS Glasgow Coma Scale, INR international normalized ratio, IQR interquartile ratio, PaCO2 partial pressure of carbon dioxide, PaO2 arterial partial oxygen pressure, pH potential hydrogen, SD standard deviation, HR heart rate, RR respiratory rate.
*: measurements are logarithmic transformed.

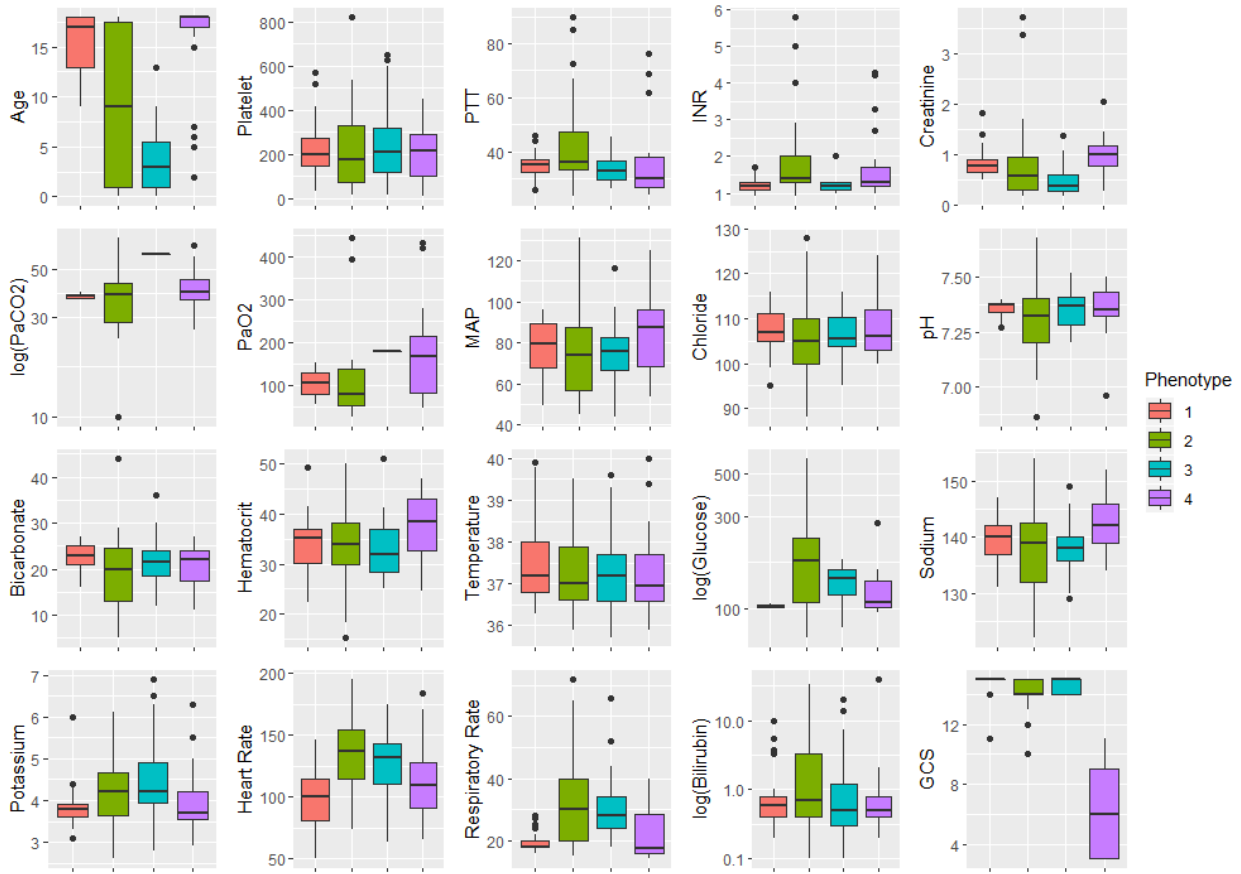

**Figure 3 Differences in clinical features by four subphenotypes**

## 3. Methods and Results: Predictive Modeling

For predictive modeling, training data was partitioned into 4 groups according to profile assignment, and predictive models were developed and evaluated within these subphenotypes. For example, for the profile 1 prediction we used 6312 non--septic cases as control data combined with the data of the 33 profile1 cases for a total of 6345 training samples. (Data from other profiles were excluded.) The profile 1 predictive performance (against an independent test set) of these profile specific sepsis models were compared with the predictive performance achieved by including all profiles in training data and predicting "any" sepsis. In effect we are comparing how accurately we can predict a sepsis subphenotype vs "any" sepsis which may be important if we can achieve higher performance prediction for a high-risk sepsis subphenotype (e.g. subphenotype 4) than for the sepsis population as a whole.

### 3.1 Training Data for Predictive Modeling

Non-septic patients of the 6446-patient cohort are used as negative controls in the predictive modeling. Features considered in the predictive model building include: 1) vital signs: heart rate, respiratory rate, body temperature, systolic blood pressure, diastolic blood pressure, GCS; 2) laboratory tests: ALT, AST, chloride, glucose, potassium, sodium, hematocrit, creatinine, PTT, INR, and platelet count; 3) blood gas measurements: pH, lactic acid, base deficit, bicarbonate, fraction of inspired oxygen, partial pressure of arterial oxygen, fraction of inspired oxygen, and partial pressure of arterial carbon dioxide. Four sets of features were extracted from data defined on different periods after admission: over first 6, 12, 24 hours following admission and over the whole length of stay. Table 3 shows simple statistics of features extracted over the whole length of stay.

| Feature | %missing | mean | SD | min | median | max |
|---|---|---|---|---|---|---|
| Age | 0.0 | 8.88 | 7.05 | 0 | 9 | 18 |
| ALT | 54.7 | 49.78 | 128.88 | 3 | 24 | 3337 |
| AST | 54.8 | 49.52 | 187.90 | 1.5 | 23 | 7794 |
| Base deficiency | 90.5 | -5.54 | 7.90 | -30 | -4 | 24 |
| Bicarbonate | 19.1 | 23.13 | 4.21 | 2 | 24 | 44 |
| Bilirubin | 54.3 | 1.16 | 3.06 | 0.05 | 0.4 | 39.3 |
| Chloride | 19.0 | 105.49 | 3.77 | 72 | 106 | 134 |
| Creatinine | 22.2 | 0.60 | 0.35 | 0.09 | 0.58 | 7.83 |
| DBP | 1.8 | 51.87 | 9.76 | 0 | 52 | 113 |
| FIO2 | 87.0 | 50.47 | 30.11 | 1 | 40 | 100 |
| GCS | 69.4 | 14.34 | 2.06 | 3 | 15 | 15 |
| Glucose | 91.2 | 181.89 | 144.83 | 6 | 118 | 600 |
| Hematocrit | 17.0 | 37.44 | 5.64 | 9 | 37.5 | 78 |
| HR | 18.3 | 116.81 | 34.13 | 1 | 114 | 265 |
| INR | 83.0 | 1.19 | 0.37 | 0.9 | 1.1 | 5.8 |
| Lactic Acid | 86.1 | 1.94 | 1.61 | 0.35 | 1.5 | 12.6 |
| PaCO2 | 90.5 | 34.36 | 10.25 | 10.9 | 33.8 | 87 |
| PaO2 | 95.4 | 165.12 | 118.21 | 23.9 | 123 | 500 |
| pH | 90.5 | 7.33 | 0.29 | 3.405 | 7.38 | 7.68 |
| Platelet | 17.9 | 291.13 | 111.44 | 1 | 274 | 1215 |
| Potassium | 19.1 | 4.15 | 0.67 | 1.2 | 4 | 8.6 |
| PTT | 89.7 | 34.62 | 17.65 | 20 | 31.2 | 200 |
| RR | 0.8 | 26.84 | 13.84 | 0 | 22 | 352 |
| SBP | 1.8 | 95.61 | 16.07 | 0 | 96 | 167 |
| Sodium | 18.9 | 138.44 | 3.45 | 54 | 139 | 163 |
| Temperature | 25.8 | 37.00 | 0.95 | 6.7 | 36.9 | 41.1 |

Abbreviations: AST aspartate aminotransferase, ALT alanine aminotransferase, PTT partial thrombin time, DBP diastolic blood pressure, SBP systolic blood pressure, GCS Glasgow Coma Scale, INR international normalized ratio, FIO2 fraction of inspired oxygen, PaCO$_2$ partial pressure of carbon dioxide, PaO$_2$ arterial partial oxygen pressure, pH potential hydrogen, SD standard deviation, HR heart rate, RR respiratory rate.

**Table 3 Basic Statistics of Training Data Features.**

## 3.2 Predictive Models

Predictive models, including gradient boosted machine (GBM) [40] and random forest (RF) [41], were built for all cases and each phenotype separately, with all non-sepsis subjects as the control group. Missing values were replaced by medians of each variables. Synthetic Minority Over-Sampling Technique (SMOTE) was applied to resample data [42]. Data was split into a training (70%) and test set (30%). Cross validation was used for tuning hyperparameters: number of trees, interaction depth, learning rate, minimum number of observations in nodes for GBM model; and number of trees for RF model. Tuned models were used to evaluate performance of predicting sepsis in the test set, and the model with best performance is selected. Confidence intervals of performance metrics were obtained by bootstrapping method [43]. To compare performance (e.g. AUC) between models, the method proposed by Delong et al. [44] was applied, with the null hypothesis that the true difference in performance metrics was equal to 0.

## 3.3 Predictive Model Results

Predictive models were built for all cases and each phenotype separately. Performances of predictive models evaluated using validation set are listed in Table 4. Model based on early 6- and 12-hour data generally yielded slightly poorer performance when comparing with models based on 24-hour and whole LOS data. Results also suggest that, providing sufficient data (e.g. 24 hours or longer stay data), significant improvements in AUC are obtained to predict phenotype 2 (p=0.0077 with 24-hour data, and p=0.0029 with whole LOS data) and phenotype 4 (p<0.0001 with 24-hour data, and p=0.0014 with whole LOS data) when comparing with the AUC from the model considering all phenotypes together. As indicated by Table 2, the two phenotypes are probably high-risk cases who used significantly more vasopressor, and phenotype 4 also had the highest mortality. Importance of features in predicting phenotypes were found to be different across phenotypes (Appendix Table A2). GCS and FIO2 ranked top in predicting phenotype 4, while age and creatinine ranked top in predicting other phenotypes.

| Metric | All Phenotypes | Phenotype 1 | Phenotype 2 | Phenotype 3 | Phenotype 4 |
|---|---|---|---|---|---|
| *Features extracted from data on the whole length of stay* | | | | | |
| AUC | 0.912(0.858 - 0.965) | 0.959(0.912 - 1.0) | 0.994(0.989 - 0.998) | 0.927(0.867 - 0.987) | 0.999(0.998 - 1.0) |
| Sensitivity | 0.925(0.8 - 1.0) | 1(0.889 - 1.0) | 1.0(1.0 - 1.0) | 0.909(0.818 - 1.0) | 1.0(1.0 - 1.0) |
| Specificity | 0.856(0.76 - 0.901) | 0.928(0.763 - 0.983) | 0.985(0.979 - 0.992) | 0.928(0.709 - 0.95) | 0.998(0.995 - 1.0) |
| Accuracy | 0.856(0.765 - 0.901) | 0.928(0.764 - 0.983) | 0.985(0.98 - 0.992) | 0.928(0.711 - 0.95) | 0.998(0.995 - 1.0) |
| PPV | 0.114(0.078 - 0.157) | 0.059(0.02 - 0.209) | 0.263(0.204 - 0.4) | 0.066(0.02 - 0.094) | 0.667(0.471 - 1.0) |
| NPV | 0.998(0.995 - 1.0) | 1(0.999 - 1.0) | 1.0(1.0 - 1.0) | 0.999(0.999 - 1.0) | 1.0(1.0 - 1.0) |
| P value* | - | 0.1947 | 0.0029 | 0.7189 | 0.0014 |
| | | | | | |
| *Features extracted from data on the first 6 hours after admission* | | | | | |
| AUC | 0.879(0.826 - 0.932) | 0.89(0.794 - 0.986) | 0.931(0.861 - 1.0) | 0.834(0.68 - 0.988) | 0.984(0.961 - 1.0) |
| Sensitivity | 0.875(0.725 - 0.95) | 0.889(0.667 - 1.0) | 0.9(0.7 - 1.0) | 0.727(0.455 - 1.0) | 1.0(1.0 - 1.0) |
| Specificity | 0.807(0.775 - 0.888) | 0.926(0.567 - 0.947) | 0.847(0.627 - 0.967) | 0.936(0.921 - 0.988) | 0.906(0.89 - 0.994) |
| Accuracy | 0.809(0.777 - 0.886) | 0.925(0.569 - 0.946) | 0.848(0.628 - 0.966) | 0.935(0.92 - 0.986) | 0.906(0.89 - 0.994) |
| PPV | 0.089(0.073 - 0.133) | 0.05(0.011 - 0.073) | 0.033(0.014 - 0.127) | 0.068(0.038 - 0.24) | 0.043(0.037 - 0.4) |
| NPV | 0.997(0.993 - 0.999) | 0.999(0.998 - 1.0) | 0.999(0.998 - 1.0) | 0.998(0.997 - 1.0) | 1.0(1.0 - 1.0) |
| P value* | - | 0.1678 | 0.1727 | 0.5119 | 0.0915 |

*Features extracted from data on the first 12 hours after admission*

| | | | | | |
|---|---|---|---|---|---|
| AUC | 0.885(0.836 - 0.933) | 0.924(0.851 - 0.998) | 0.9(0.764 - 1.0) | 0.895(0.816 - 0.975) | 0.989(0.974 - 1.0) |
| sensitivity | 0.875(0.675 - 1.0) | 0.889(0.778 - 1.0) | 0.9(0.6 - 1.0) | 0.909(0.727 - 1.0) | 1.0(1.0 - 1.0) |
| specificity | 0.772(0.624 - 0.931) | 0.964(0.682 - 0.977) | 0.871(0.847 - 0.999) | 0.806(0.552 - 0.948) | 0.943(0.93 - 1.0) |
| accuracy | 0.775(0.631 - 0.927) | 0.963(0.683 - 0.976) | 0.871(0.848 - 0.998) | 0.806(0.554 - 0.947) | 0.944(0.931 - 1.0) |
| PPV | 0.076(0.05 - 0.183) | 0.1(0.015 - 0.153) | 0.037(0.027 - 0.769) | 0.028(0.013 - 0.082) | 0.07(0.057 - 1.0) |
| NPV | 0.996(0.993 - 1.0) | 0.999(0.999 - 1.0) | 0.999(0.998 - 1.0) | 0.999(0.998 - 1.0) | 1.0(1.0 - 1.0) |
| P value* | - | 0.3794 | 0.8358 | 0.831 | <0.0001 |
| ***Features extracted from data on the first 24 hours after admission*** | | | | | |
| AUC | 0.918(0.881 - 0.956) | 0.922(0.851 - 0.994) | 0.976(0.956 - 0.997) | 0.944(0.893 - 0.995) | 0.998(0.996 - 1.0) |
| sensitivity | 0.85(0.725 - 0.95) | 0.889(0.778 - 1.0) | 1.0(1.0 - 1.0) | 0.909(0.818 - 1.0) | 1.0(1.0 - 1.0) |
| specificity | 0.899(0.757 - 0.928) | 0.882(0.646 - 0.971) | 0.924(0.911 - 0.978) | 0.904(0.697 - 0.964) | 0.994(0.989 - 0.999) |
| accuracy | 0.898(0.759 - 0.926) | 0.882(0.647 - 0.971) | 0.924(0.911 - 0.978) | 0.904(0.699 - 0.964) | 0.994(0.989 - 0.999) |
| PPV | 0.149(0.073 - 0.195) | 0.036(0.013 - 0.127) | 0.065(0.056 - 0.192) | 0.054(0.019 - 0.133) | 0.421(0.276 - 0.889) |
| NPV | 0.996(0.994 - 0.999) | 0.999(0.999 - 1.0) | 1.0(1.0 - 1.0) | 0.999(0.999 - 1.0) | 1.0(1.0 - 1.0) |
| P value* | - | 0.9217 | 0.0077 | 0.4215 | <0.0001 |

*: P value is from AUC comparisons, in which "All phenotypes" is the reference AUC.

**Table 4 Comparative performance between subphenotype-specific and all sepsis phenotype prediction**

## 4. Discussion

Using routinely available clinical variables this study applied LPA clustering to group pediatric sepsis patients to identify four latent profiles (i.e. homogeneous subphenotypes) and derived separate predictive models for each subphenotype. The predictive models performed better on specific subphenotypes of sepsis patients than on the sepsis cohort overall, thus supporting our primary hypothesis. Further, we found that, prediction was better in two specific subphenotype groups than in the other subphenotypes. Profile 4 patients were characterized by neurological dysfunction (lowest GCS total score) and highest mortality rate (22.2%); profile 2 characterized by respiratory dysfunction (low PaO2). Patients with profiles 2 and 4 had significant more vasopressor use, while profile 2 patients had lowest BP. Although our study did not focus on this issue, given ongoing concerns regarding the use of aggressive fluid management in some patients [45], the identification of these subphenotypes may help triage pediatric sepsis patients that respond differently to aggressive fluids treatment [32][46].

At an AUC of .98 compared to AUC of .88 (p < .0001) for Profile 4 using data within 12 hours following admission, our results suggest that significantly improved performance of prediction can be obtained for pediatric sepsis subphenotypes at high risk of mortality. Although there are significant differences in physiology, underlying conditions, mortality rates and even prevalence of sepsis between pediatric and adults, a recent related study of adult sepsis also found four sepsis phenotypes with different demographics, laboratory values, and patterns of organ dysfunction, with treatment outcomes sensitive to changes in the distribution of these subphenotypes [47].

There are a number of ways to perform agnostic clustering, such as hierarchical cluster analysis, self-organizing map, K-means consensus clustering [48], and latent class analysis (LCA) [28], [30], [49]. Latent class analysis (LCA) and latent profile analysis (LPA) are techniques that aim to recover hidden groups from observed data. They are similar to clustering techniques but more flexible because they are based on an explicit model of the data and allow you to account for the fact that the recovered groups are uncertain by deriving clusters using a probabilistic model that describes distribution of the data. So instead of looking for clusters with some arbitrary chosen distance measure, LPA fits a model that describes distribution of the data and based on this model you assess probabilities that certain patients are members of certain latent profiles. LCA and LPA are useful when you want to reduce many continuous (LPA) or categorical (LCA) variables to a few subgroups. They can also help experimenters in situations where the treatment effect is different for subgroups, but the subgroups have not been identified [50]. For sepsis, LPA seems to be an appropriate clustering choice given the continuous variables underlying organ dysfunction and since it allows the modeling of a latent structure underlying the observed data (e.g. "dysregulated host response") rather than just modeling similarities based on a distance metric (e.g. k-means clustering) [32].

It is known that heterogeneity in population poses a great challenge to predictive modeling. First, the training data may be comprised of instances from not just one distribution, but several distributions juxtaposed together. In the presence of multimodality within the classes, there may be imbalance among the distribution of different modes in the training set. Hence, some of the modes may be underrepresented during training, resulting in poor performance on those modes during the testing stage. Second, while some of the modes of a particular class may be easy to distinguish from modes of the other class, there may be modes that participate in class confusion, i.e., reside in regions of feature space that overlap with instances from other classes. The presence of such overlapping modes can degrade the learning of any classification model trained across all modes of every class. Third, even if we are able to learn a predictive model that shows reasonable performance on the training set, the test set may have a completely different distribution of data instances than the training set, as the populations of training and test sets can be different. Hence, the training performance can be quite misleading as it may not always be reflective of the performance on test instances. For these reasons, identifying homogeneous subgroups within heterogeneous population mitigate the impact of population's heterogeneity in predictive model building. Significant differences in importance of features across subphenotype predictive models (e.g. subphenotypes 2 and 4 as seen in Table A3 in the Appendix), compared to the entire cohort support this line of reasoning.

Several limitations must be acknowledged in this study. First, this study used EHR data which were produced by routine clinical practice with significant missing data, although in many cases missing data may not be at random and training feature "missingness" may reflect clinical decisions that can be modeled [51] . As described above imputations were performed for variables containing missing values. Although there are many sophisticated methods to deal with missing values, significant bias may be introduced for those with missing rates greater than 40% [52]. For example, in this study the number of missing serum lactate observations was high while studies show that in children treated for sepsis in the emergency department, lactate levels greater than 36 mg/dL (4 mmol/L) were associated with mortality, but also had a low sensitivity, potentially explaining the level of "missingness" [53].  While its use is not endorsed in pediatric sepsis guidelines [54], the measurement of lactate levels may have utility in early risk stratification of pediatric sepsis [53]. Second, although restricting the variables used for modeling to those available in clinical practice is reasonable, it may limit the separation of classes. It would be better to use biomarkers and genomics as well, since studies indicate they may contribute in pediatric sepsis subphenotyping [55]. Combining clinical data and biological data in LPA-based phenotyping may improve homogeneities of sepsis sub-phenotypes towards further enhancement of predictive performance of ML within subgroups. However, inflammatory or genetic biomarker biomarkers were not routinely obtained and therefore were not available in the MedStar database. Finally, the study was based on a relatively small sample size and requires external validation with other datasets.

## 5. Conclusions
Rapid protocolized treatment is known to improve sepsis outcomes, however especially in children, early diagnosis remains challenging due to age-dependent heterogeneity and complex presentation [56]. Moreover, there is growing recognition of the presence of sepsis subphenotypes identified by profiling clinical data that respond differently to treatment [47].  While promising machine learning models derived over large samples have the potential to accurately identify children with sepsis hours before clinical recognition [56], the potential benefits of incorporating sepsis subphenotype knowledge in training data is unknown. This study utilized LPA to identify four clinically meaningful pediatric sepsis subphenotypes in training data and, for two of these subphenotypes with the highest mortality, derived enhanced predictive models on the partitioned data. These experiments suggest that LPA is an applicable tool useful in analyzing heterogeneous pediatric sepsis to identify subphenotypes beneficial for building enhanced predictive pediatric sepsis models. Additional studies with larger data samples needed to validate our findings.

## 6. Declarations

**6.1 Ethics approval and consent to participate.**

Based on NIH research study protocols: *Protection of Human Subjects* and associated Research Plan: *Rule-based Semantics and Big Data Based Methods for Effective Clinical Decision Support (CDS): A Pediatric Severe Sepsis Case Study* submitted to the Advarra IRB with reference Pro00021165, the IRB reviewed the project in accordance with the 45 CFR Part 46, Subpart D Federal Regulations which provide for additional protections for children as research subjects and determined that the research study meets the criteria found in the risk category described as follows:

☐ 45 CFR 46.404: *"Research not involving greater than minimal risk."*

Approval with a Waiver of Parental Consent and Assent was obtained April 15, 2017 and recently continued April 25, 2019.

**6.2 Consent for publication**. Not applicable.

**6.3 Availability of data and materials.** The data that support the findings of this study, performed under a NIH SBIR grant (see **Funding** below), may be available from MedStar Health Research Institute (MHRI), but restrictions apply to the availability of these data, which were made available under a data sharing/Business Associate Agreement and with IRB review. In general, MHRI medical record data are not publicly available. Study data however may be available from the authors upon reasonable request and with MHRI permission.

**6.4 Competing Interests**. This study was partially funded as a NIH SBIR awarded to CTA. NIH SBIRs have the objective of translating promising technologies to the private sector and enable life-saving commercialized innovations to reach consumer markets, e.g. hospitals, urgent care centers, clinics, etc. As employees of, or consultants/advisors to CTA, all authors understand the commercialization goals of SBIRs and therefore to that extent, have competing financial and/or non-financial interests in the success of this technology.

**6.5 Funding.** Research reported in this publication was supported by a NIH SBIR award to CTA by NIH National Institute of General Medical Sciences under award number 1R43GM122154 – 01. Additional analysis following the SBIR grant period of performance and development of this manuscript funded by CTA.

**6.6 Authors' contributions.** TV performed as the Principal Investigator/Program Director of the underlying NIH research study, conceived the current study, and authored major segments of the **Introduction, Discussion and Conclusions** sections of the manuscript. TV is the guarantor of the article, taking responsibility for the integrity of the work, from inception to published article. TW performed as the principal LPA/machine learning analyst and authored major sections of **Methods and Results** sections of the manuscript. TT provided the technical infrastructure for the MedStar data used in the study used for LPA and predictive analytics as well as the rule engine used to detect onset time of sepsis. EA provided data management services associated with training data preparation. Pediatric clinicians IK, JC provided clinical support for the logic used to detect pediatric sepsis (using concept maps such as illustrated in Fig. 1) as well as the clinical interpretation of the results of the LPA analysis Table 2). Sepsis clinicians UA and SY provided clinical support in interpretation of results as compared to adult sepsis, including guidance on features used for training data. All authors read and approved the manuscript.

**6.7 Acknowledgements.** Not applicable.

**Abbreviations**
ALT: Alanine aminotransferase; ARDS: Acute Respiratory Distress Syndrome; AST: Aspartate aminotransferase; AUC: Area under receiver operating curve; BIC: Bayesian information criteria; BP: Blood pressure; CI: Confidence interval; DBP: Diastolic Blood Pressure; EHR: Electronic Health Record; EM: Expectation Maximization; GBM: Gradient Boosted Machine; GCS: Glasgow Coma Scale; HR: Heart Rate; ICD: International Classification of Diseases; ICU: Intensive care unit; INR: International normalized ratio; IQR: Interquartile range; IRB: Institutional review board; LCA: Latent class analysis; LOS: Length of stay; LPA: Latent profile analysis; MAP: Mean Arterial Pressure; ML: machine Learning; NPV: Negative Predictive Value; OD: Organ Dysfunction; PaCO2: Partial pressure of carbon dioxide; PaO2: Arterial partial oxygen pressure; pH: potential of Hydrogen; PICU: Pediatric Intensive Care Unit; PPV: Positive Predictive

Value; PTT: Partial thrombin time; RF: Random Forest; ROC: receiver operating curve; RR: Respiratory rate; SBP: Systolic Blood Pressure; SD: Standard Deviation; SIRS: Systemic Inflammatory Response Syndrome; SOFA: Sequential organ failure assessment; VEI = diagonal, varying volume, equal shape.

# Appendix

## Table A1 Identifiers used in Fig.2

| Identifier | Distribution | Volume | Shape | Orientation |
|---|---|---|---|---|
| EII | Spherical | equal | equal | NA |
| VII | Spherical | variable | equal | NA |
| EEI | Diagonal | equal | equal | coordinate axes |
| VEI | Diagonal | variable | equal | coordinate axes |
| EVI | Diagonal | equal | variable | coordinate axes |
| VVI | Diagonal | variable | variable | coordinate axes |
| EEE | Ellipsoidal | equal | equal | equal |
| EEV | Ellipsoidal | equal | equal | variable |
| VEV | Ellipsoidal | variable | equal | variable |
| VVV | Ellipsoidal | variable | variable | variable |

## Table A2 Variable Importance in predicting phenotypes

| Importance rank | All phenotypes Variable | Score | Phenotype 1 Variable | Score | Phenotype 2 Variable | Score | Phenotype 3 Variable | Score | Phenotype 4 Variable | Score |
|---|---|---|---|---|---|---|---|---|---|---|
| 1 | Age | 100 | Age | 100 | Age | 100 | Age | 100 | GCS | 100 |
| 2 | Creatinine | 66.93 | Creatinine | 63.4 | RR | 58.23 | Creatinine | 53.37 | FIO2 | 45.64 |
| 3 | INR | 29.16 | Hematocrit | 51.7 | Base Deficiency | 29.76 | RR | 31.44 | Lactic Acid | 40.37 |
| 4 | Base Deficiency | 28.13 | INR | 51.31 | PaO2 | 27.43 | INR | 24.11 | Creatinine | 15.4 |
| 5 | pH | 27.21 | RR | 48.41 | HR | 20.75 | ALT | 19.19 | INR | 13.67 |
| 6 | Bicarbonate | 26.81 | Lactic Acid | 41.77 | pH | 19.73 | HR | 17.99 | PaO2 | 13.57 |
| 7 | AST | 24.98 | PTT | 39.51 | PaCO2 | 17.49 | GCS | 16.15 | Sodium | 7.48 |
| 8 | Lactic Acid | 21.73 | Platelet | 39.45 | INR | 17.42 | pH | 14.99 | PTT | 6.49 |
| 9 | RR | 19.03 | Temperature | 32.58 | FIO2 | 16.79 | AST | 14.08 | Age | 4.49 |
| 10 | Chloride | 18.97 | HR | 27.37 | Bicarbonate | 15.51 | Potassium | 12.3 | Bilirubin | 3.99 |
| 11 | Hematocrit | 18.38 | DBP | 26.35 | Bilirubin | 13.86 | Lactic Acid | 11.6 | AST | 1.68 |
| 12 | ALT | 15.94 | Sodium | 26.33 | Creatinine | 12.75 | Base Deficiency | 11.31 | DBP | 1.67 |
| 15 | PTT | 12.76 | AST | 23.72 | AST | 11.56 | Bicarbonate | 9.78 | Base Deficiency | 1.57 |
| 16 | PaCO2 | 12.44 | PaCO2 | 23.32 | Temperature | 11 | PTT | 9.75 | Chloride | 1.38 |
| 17 | Bilirubin | 12.14 | Chloride | 22.7 | Hematocrit | 9.93 | Temperature | 7.32 | Platelet | 1.2 |
| 18 | HR | 10.82 | SBP | 21.52 | DBP | 9.14 | PaCO2 | 7.05 | Potassium | 1.05 |
| 19 | SBP | 10.59 | Potassium | 18.99 | Potassium | 9.03 | Hematocrit | 6.74 | RR | 0.66 |
| 22 | FIO2 | 7.09 | Bilirubin | 12.71 | ALT | 7.51 | Sodium | 4.89 | SBP | 0.56 |
| 23 | Glucose | 7.01 | pH | 9.91 | Sodium | 7.18 | PaO2 | 4.71 | Bicarbonate | 0.53 |
| 24 | Sodium | 3.25 | FIO2 | 1.97 | PTT | 6.58 | DBP | 3.75 | HR | 0.31 |
| 25 | Potassium | 1.12 | PaO2 | 1.49 | Platelet | 5.61 | SBP | 3.27 | Glucose | 0.16 |
| 26 | Temperature | 0 | Glucose | 0 | Lactic Acid | 0 | FIO2 | 0 | Temperature | 0 |

**Table A3 ICD-9 Codes for Infection**

ICD-9 codes for infection:

```sql
WITH infection_group AS
(
    SELECT subject_id, hadm_id,
    CASE
        WHEN substring(icd9_code,1,3) IN
('001','002','003','004','005','008',

'009','010','011','012','013','014','015','016','017','018',

'020','021','022','023','024','025','026','027','030','031',

'032','033','034','035','036','037','038','039','040','041',

'090','091','092','093','094','095','096','097','098','100',

'101','102','103','104','110','111','112','114','115','116',

'117','118','320','322','324','325','420','421','451','461',

'462','463','464','465','481','482','485','486','494','510',

'513','540','541','542','566','567','590','597','601','614',
                '615','616','681','682','683','686','730') THEN 1
        WHEN substring(icd9_code,1,4) IN
('5695','5720','5721','5750','5990','7110',
                        '7907','9966','9985','9993') THEN 1
        WHEN substring(icd9_code,1,5) IN
('49121','56201','56203','56211','56213',
                        '56983') THEN 1
        ELSE 0 END AS infection
    FROM diagnoses_icd
```